\documentclass[12pt,onecolumn]{IEEEtran}
\usepackage{epsfig}
\usepackage{graphicx}
\usepackage{amsmath}
\usepackage{amssymb}
\usepackage{enumerate}
\usepackage{algorithm}
\usepackage{algorithmic}
\usepackage{amsthm}
\usepackage{subfigure}
\usepackage{epstopdf}

\newtheorem{theorem}{Theorem}[]

\DeclareMathOperator*{\argmin}{arg\,min}

\begin{document}
%
% paper title
% Titles are generally capitalized except for words such as a, an, and, as,
% at, but, by, for, in, nor, of, on, or, the, to and up, which are usually
% not capitalized unless they are the first or last word of the title.
% Linebreaks \\ can be used within to get better formatting as desired.
% Do not put math or special symbols in the title.
\title{Robust Subspace Clustering via Smoothed Rank Approximation}
%
%
% author names and IEEE memberships
% note positions of commas and nonbreaking spaces ( ~ ) LaTeX will not break
% a structure at a ~ so this keeps an author's name from being broken across
% two lines.
% use \thanks{} to gain access to the first footnote area
% a separate \thanks must be used for each paragraph as LaTeX2e's \thanks
% was not built to handle multiple paragraphs
%

\author{Zhao~Kang,
        Chong~Peng,
        and~Qiang~Cheng$^\ast$% <-this % stops a space
%\thanks{Copyright (c) 2015 IEEE. Personal use of this material is permitted. However, permission to use this material for any other purposes must be obtained from the IEEE by sending a request to pubs-permissions@ieee.org.}
\thanks{$^\ast$The authors are with Computer Science Department, Southern Illinois University, Carbondale, IL, USA, 62901. Correspondence should be sent to qcheng@cs.siu.edu.}}% <-this % stops a space

\maketitle

% As a general rule, do not put math, special symbols or citations
% in the abstract or keywords.
\begin{abstract}
Matrix rank minimizing subject to affine constraints arises in many application areas, ranging from  signal processing to machine learning. Nuclear norm is a convex relaxation for this problem which can recover the rank exactly under some restricted and theoretically interesting conditions. However, for many real-world applications, nuclear norm approximation to the rank function can only produce a result far from the optimum. To seek a solution of higher accuracy than the nuclear norm, in this paper, we propose a rank approximation based on Logarithm-Determinant. We consider using this rank approximation for subspace clustering application. Our framework can model different kinds of errors and noise. Effective optimization strategy is developed with theoretical guarantee to converge to a stationary point. The proposed method gives promising results on face clustering and motion segmentation tasks compared to the state-of-the-art subspace clustering algorithms.
\end{abstract}

% Note that keywords are not normally used for peerreview papers.

% For peer review papers, you can put extra information on the cover
% page as needed:
% \ifCLASSOPTIONpeerreview
% \begin{center} \bfseries EDICS Category: 3-BBND \end{center}
% \fi
%
% For peerreview papers, this IEEEtran command inserts a page break and
% creates the second title. It will be ignored for other modes.
%\IEEEpeerreviewmaketitle

\section{Introduction}
% The very first letter is a 2 line initial drop letter followed
% by the rest of the first word in caps.
% 
% form to use if the first word consists of a single letter:
% \IEEEPARstart{A}{demo} file is ....
% 
% form to use if you need the single drop letter followed by
% normal text (unknown if ever used by IEEE):
% \IEEEPARstart{A}{}demo file is ....
% 
% Some journals put the first two words in caps:
% \IEEEPARstart{T}{his demo} file is ....
% 
% Here we have the typical use of a "T" for an initial drop letter
% and "HIS" in caps to complete the first word.
\IEEEPARstart{R}{ecently} there has been a surge of interest in finding minimum rank matrix within an affine constraint set \cite{candes2009exact,liu2010robust}%,liu2013robust},lu2014generalized}. 
. The problem is as follows,
\begin{equation}
\label{affine}
\min_Z \quad rank(Z)\quad s.t\quad  \mathbf{\mathcal{A}}(Z)=b,
\end{equation}
where $Z\in \mathbf{\mathcal{R}}^{m\times n}$ is the unknown matrix, $\mathbf{\mathcal{A}}:\mathbf{\mathcal{R}}^{m\times n}\rightarrow \mathbf{\mathcal{R}}^p$ is a linear mapping, and $b\in \mathbf{\mathcal{R}}^p $ denotes the observations. Unfortunately, however, minimizing the rank of a matrix is known to be NP-hard and a very challenging problem. 

Consequently, a widely-used convex relaxation approach is to replace the rank function with the nuclear norm $\|Z\|_*=\sum_{i=1}^{n} \sigma_i(Z)$, where $\sigma_i(Z)$ is the $i$-th singular value of $Z$ (suppose $n<m$). The nuclear norm technique has been shown to be effective in encouraging a low-rank solution \cite{fazel2002matrix,recht2010guaranteed}. Nevertheless,  there is no guarantee for the minimum nuclear norm solution to coincide with that of minimal rank in many interesting circumstances, which is heavily dependent on the singular values of matrices in the nullspace of $\mathcal{A}$. Some variations of the nuclear norm have demonstrated promising results, e.g., singular value thresholding \cite{cai2010singular}, and truncated nuclear norm \cite{hu2013fast}.  

Another popular alternative approach is to compute 
\begin{equation}
\min_Z \sum_{i=1}^n h(\sigma_i(Z)) \quad s.t. \quad \mathbf{\mathcal{A}}(Z)=b,
\end{equation}
where $h$ is usually a nonconvex and nonsmooth function. It has been observed that nonconvex approach can succeed in a broader range of scenarios \cite{lu2014generalized}. However, nonconvex optimization is often challenging.%, especially on potentially large-scale data. %mohan2012generalized, 

To overcome the above-mentioned difficulties, in this paper, we propose to use a particular  log-determinant (LogDet) function to approximate the rank function. The formulation we consider is:
\begin{equation}
F(Z)=logdet(I+Z^TZ)=\sum_{i=1}^n  log(1+\sigma_i^2(Z)),
\label{logform}
\end{equation}
where $I\in\mathbf{\mathcal{R}}^{n\times n}$ is an identity matrix. For large nonzero singular values, the LogDet function value will be much smaller than the nuclear norm. It is easy to show that $logdet (I + Z^T Z) \le \left\|Z\right\|_*$. Therefore, LogDet is a tighter rank approximation function than the nuclear norm. Although a similar function $logdet(Z+\delta I)$ has been proposed and iterative linearization has been used to find a local minimum \cite{fazel2003log}, its $\delta$ is required to be small (e.g., $10^{-6}$), which leads to significantly biased approximation for small singular values and thus limited applications. Smoothed Schatten-$p$ function, $Tr(Z^TZ+\gamma I)^{p/2}$, has been well studied in matrix completion \cite{mohan2012iterative}; nonetheless, the resulting algorithm is rather sensitive to parameter $\gamma$.   

The main contributions of this work are as follows: $1)$ An efficient algorithm is devised to optimize LogDet associated nonconvex objective function; $2)$ Our method pushes the accuracy of subspace clustering to a new level. 
\section{Problem statement of subspace clustering}
An important application of the proposed LogDet function is the low-rank representation based subspace clustering problem. There has been significant research effort on this subject over
the past several years due to its promising applications in computer vision and machine learning  \cite{vidal2010tutorial}. Subspace clustering aims at finding a low-dimensional subspace for each group of points, which is based on the widely-used assumption that high-dimensional data actually reside in a union of multiple low-dimensional subspaces. Under such an assumption the data could be separated in a projected subspace. Consequently, subspace clustering mainly involves two tasks, firstly projecting the data into a latent subspace to describe the affinities of points, and subsequently, grouping the data in that subspace. Some spectral clustering methods %\cite{ng2002spectral,von2007tutorial} 
such as normalized cuts (NCuts) \cite{shi2000normalized}  are usually used in the second task to find  the cluster membership. Besides this spectral clustering-based subspace clustering method, iterative, algebraic, and statistical methods are also available in the literature \cite{vidal2010tutorial}, but they are usually sensitive to initialization, noise or outliers.

Typical spectral clustering-based subspace clustering methods are Local Subspace Affinity (LSA) \cite{yan2006general}, Sparse Subspace Clustering (SSC) \cite{elhamifar2013sparse}, Low Rank Representation (LRR) \cite{liu2010robust,liu2013robust} and its more robust variant LRSC \cite{favaro2011closed,vidal2014low}. Among them, SSC and LRR give promising results even in the presence of large outliers or corruption \cite{liu2012exact,wang2013noisy}. They both suppose that each data point can be written as a linear combination of other points in the dataset. SSC tries to find the sparsest representation of data points through $l_1$-norm. Even when the subspaces overlap, SSC can successfully reveal subspace structure \cite{soltanolkotabi2012geometric}. SSC's solution is sometimes too sparse to form a fully connected affinity graph for data in a single subspace \cite{nasihatkon2011graph}. LRR uses the lowest-rank representation to depict the similarity among data points. It is theoretically guaranteed to succeed when the subspaces are independent. 

Let $X=[x_1, x_2, \cdots, x_n]\in \mathbf{\mathcal{R}}^{m\times n}$ store a set of $n$ $m$-dimensional samples drawn from a union of $k$ subspaces. LRR considers the following regularized nuclear norm rank minimization problem:
\begin{equation}
\min_{Z, E} \|Z\|_*+\lambda \|E\|_l \quad s.t. \quad X=XZ+E,
\end{equation}
where $\lambda>0$ is a parameter, $E$ represents unknown corruption, and $\| \cdot \|_l$ can be $l_{2,1}$-norm, $l_{1}$-norm, or squared Frobenius norm. Specifically, if random corruption is assumed in the data, $\|E\|_1:=\sum_{i=1}^m \sum_{j=1}^n |E_{ij}|$ is usually adopted; $\|E\|_{2,1}:=\sum_{j=1}^n \sqrt{\sum_{i=1}^m E_{ij}^2}$ is more suitable to characterize sample-specific corruptions and outliers; $\|E\|_F^2:=\sum_{i=1}^m \sum_{j=1}^n E_{ij}^2$ often describes Gaussian noise. LRR is able to produce pretty competitive performance on subspace clustering in the current literature. However, the solution to it might not be unique due to the nuclear norm \cite{zhang2013counterexample}; and furthermore, the rank surrogate can deviate far from the true rank function.  

To better approximate the rank while possessing the desired robustness similar to LRR, in this paper, we propose to use the above-mentioned LogDet function and solve the following problem:
\begin{equation}
\min_{Z, E} logdet(I+Z^TZ)+\lambda \|E\|_l \quad s.t. \quad X=XZ+E.
\label{ourproblem}
\end{equation}
The objective function of (\ref{ourproblem}) is nonconvex. We design an effective optimization strategy based on an augmented Lagrangian multiplier (ALM) method, which is potentially applicable to large-scale data because of the decomposability of ALM and its admittance to parallel algorithms. For our optimization method, we provide theoretical analysis for its convergence, which mathematically guarantees that our algorithm can produce a convergent subsequence and the converged point is a stationary point of (\ref{ourproblem}). 

\section{Proposed method: CLAR}
In this section, we present the proposed robust subspace clustering algorithm CLAR: Clustering with Log-determinant Approximation to Rank. The basic theorems and optimization algorithm will be presented below.

\subsection{Smoothed rank minimization}
To make the objective function in (\ref{ourproblem}) separable, we first convert it to the following equivalent problem by introducing an auxiliary variable $J$: 
\begin{equation}
\label{prob}
\min_{E, J, Z}\hspace{.1cm} logdet(I+J^TJ)+\lambda ||E||_l\hspace{.1cm}s.t. \hspace{.1cm}X=XZ+E, Z=J.
\end{equation}
We can solve problem (\ref{prob}) using a type of ALM method. The corresponding augmented Lagrangian function is 
\begin{equation}
\begin{split}
&L(E, J, Y_1, Y_2, Z, \mu) =logdet(I+J^TJ)+\lambda \|E\|_l  \\
&+Tr(Y_1^T(J-Z)) +Tr(Y_2^T(X-XZ-E))+\\
&\frac{\mu}{2}(\|J-Z\|_F^2+\|X-XZ-E\|_F^2),
\end{split}
\end{equation}
where $Y_1$ and $Y_2$ are Lagrange multipliers, and $\mu>0$ is a penalty parameter. Then we can apply the alternating minimization idea to update  one of the variables with the others fixed.  

Given the current point $E^t$, $J^t$, $Z^t$, $Y_1^t$, $Y_2^t$, the updating scheme is: 
\begin{eqnarray*}
\begin{split}
Z^{t+1}&=\argmin_Z\hspace{.01cm}\hspace{.01cm}Tr[(Y_1^t)^T(J^{t}-Z)]+\frac{\mu^t}{2}\|J^{t}-Z\|_F^2\\
&+Tr[(Y_2^t)^T(X-XZ-E^t)]+\frac{\mu^t}{2}\|X-XZ-E^t\|_F^2, \\
J^{t+1}&= \argmin_J \hspace{.01cm}log\hspace{.01cm} det(I+J^TJ)+\\
&\frac{\mu^t}{2}\|J-(Z^{t+1}-\frac{Y_1^t}{\mu^t})\|_F^2,\\
E^{t+1}&=\argmin_E \hspace{.01cm}\lambda ||E||_l+Tr[(Y_2^{t})^T(X-XZ^{t+1}-E)]\\
&+\frac{\mu^t}{2}||X-XZ^{t+1}-E||_F^2. 
\end{split}
\end{eqnarray*}
The first equation above has a closed-form solution:
\begin{equation}
\label{solveZ}
Z^{t+1}=(I+X^TX)^{-1}[X^T(X-E^t)+J^{t}+\frac{Y_1^t+X^TY_2^t}{\mu^t}].
\end{equation}
%For $J^{t+1}$: 
%\begin{equation}
%J^{t+1}= \min_J \hspace{.1cm}log\hspace{.2cm} det(I+J^TJ)+\frac{\mu}{2}||J-(Z^t-\frac{Y_1^t}{\mu})||_F^2.
%\end{equation}
For $J$ updating, it can be converted to scalar minimization problems due to the following theorem \cite{kang2015logdet}, which is also proved in
the supplementary material.  
\begin{theorem}
If $F(Z)$ is a unitarily invariant function and SVD of $A$ is $A = U \Sigma_A V^T$, then the optimal solution to the following problem 
\begin{equation}
\min_{Z}F(Z)+\frac{\beta}{2}\|Z-A\|_F^{2}
\label{eq:Zobj}
\end{equation}
 is $Z^*$ with SVD $U\Sigma_Z^* V^T$, where $\Sigma_Z^* = diag\left(\sigma^*\right)$; moreover, $F(Z) = f \circ \sigma(Z)$, where $\sigma(Z)$ is the vector of nonincreasing  singular values of $Z$, % with $f$ being an absolutely symmetric function\footnote{A function $f(\cdot)$ is absolutely symmetric if $f(x)$ is invariant under arbitrary permutations and sign changes of the components of $x$ \cite{lewis1995convex}.}, 
 then $\sigma^*$ is obtained by using the Moreau-Yosida proximity operator
$\sigma^* = prox_{f, \beta} (\sigma_{A})$, where $\sigma_A := diag(\Sigma_A)$, and 
\begin{equation}
\label{scalar}
prox_{f, \beta} (\sigma_A) := \argmin_{\sigma} f(\sigma) + \frac{\beta}{2}\|\sigma - \sigma_A\|_2^2.
\end{equation}
\end{theorem}
%\begin{figure}[!t]
%\centering
%\includegraphics[width=2.5in]{myfigure}
% where an .eps filename suffix will be assumed under latex, 
% and a .pdf suffix will be assumed for pdflatex; or what has been declared
% via \DeclareGraphicsExtensions.
%\caption{Simulation results for the network.}
%\label{fig_sim}
%\end{figure}

According to the first-order optimality condition, the gradient of the objective function of  (\ref{scalar}) with respect to each singular value should vanish.  
Thus we have 
\begin{equation}
\label{svdf}
\frac{2\sigma_i}{1+\sigma_{i}^{2}}+\beta_k (\sigma_{i}-\sigma_{i,A}^{t})=0,\hspace{0.1cm} s.t. \hspace{0.1cm}\sigma_i \ge  0, \hspace{.1cm} for\hspace{0.1cm} i=1, ..., n.\\
\end{equation} 
The above equation is cubic and gives three roots. If $\sigma_{i,A}^t=0$, the minimizer $\sigma_i^*$ will be 0; otherwise, it can be shown that there is a unique minimizer $\sigma_i^{*}\in [0, \sigma_{i,A}^{t})$ if $\beta>\frac{1}{4}$. To ensure this requirement is satisfied, we adopt $\mu^0=0.4$ in our experiments. Finally, we obtain the update of $J$ variable with
\begin{equation}
\label{solveJ}
J^{t+1}=U diag(\sigma_{1}^{t+1}, ..., \sigma_{n}^{t+1}) V^T.
\end{equation}
Depending on different regularization strategies, we have different closed-form solutions for $E$. 
For squared Forbenius norm, 
\begin{equation}
\label{error2}
E^{t+1}=\frac{Y_2^{t}+\mu^t(X-XZ^{t+1})}{\mu^t+2\lambda}.
\end{equation}
For $l_1$-norm, %we can use lemma \ref{lemma1} 
according to \cite{beck2009fast},
%\begin{lemma}
%\label{lemma1}
%For $\mu>0$, and $K\in\mathbf{\mathcal{R}}^{s\times t}$, the solution of the problem  
%\begin{equation*}
%\min_L\vspace{.2cm} \mu\|L\|_1+\frac{1}{2}\|L-K\|_F^2, 
%\end{equation*}
%is given by $L_\mu(K)$, which is defined component-wisely by
%\begin{equation*}
%[L_{\mu}(K)]_{ij}=max\{|K_{ij}|-\mu,0\}\cdot sign(K_{ij}).
%\end{equation*}
%\end{lemma}
if we define $Q=X-XZ^{t+1}+\frac{Y_1^t}{\mu^t}$, then $E$ can be updated element-wisely as:  
\begin{eqnarray}
\label{error1}
E_{ij}^{t+1} = \left\{
\begin{array}{ll} Q_{ij}-\frac{\lambda}{\mu^t}sgn(Q_{ij}) , 
&\mbox{if $ |Q_{ij}| <\frac{\lambda}{\mu^t}$};\\
0, &\mbox{otherwise.}
\end{array}\right. 
\end{eqnarray}
For $l_{2,1}$-norm, %we can use lemma \ref{lemma2} \cite{yang2009fast}.
%\begin{lemma} \label{lemma2} 
%Let $H$ be known. If the optimal solution to
%\begin{eqnarray*}
%\min_{W} \alpha\|W\|_{2,1} + \frac{1}{2}\|W-H\|_F^2
%\end{eqnarray*}
%is $W^*$, then the $i$-th column of $W^*$ is
%\begin{eqnarray*}
%[W^*]_{:,i}=\left\{
%\begin{array}{ll} \frac{\|H_{:,i}\|_2-\alpha}{\|H_{:,i}\|_2}H_{:,i}, & \mbox{if $\|H_{:,i}\|_2>\alpha$};\\
%0, & \mbox{otherwise.}
%\end{array}\right.
%\end{eqnarray*}
%\end{lemma}
by \cite{yang2009fast}, we have
\begin{eqnarray}
\label{error21}
[E^{t+1}]_{:,i}=\left\{
\begin{array}{ll} \frac{\left\|Q_{:,i}\right\|_2-\frac{\lambda}{\mu^t}}{\left\|Q_{:,i}\right\|_2}Q_{:,i}, & \mbox{if $\left\|Q_{:,i}\right\|_2>\frac{\lambda}{\mu^t}$};\\
0, & \mbox{otherwise.}
\end{array}\right.
\end{eqnarray}
\begin{algorithm}[tb]
\tiny
   \caption{Smoothed Rank Minimization}
   \label{alg:rankminimization}
  {\bfseries Input:} data matrix $X\in \mathbf{\mathcal{R}}^{m\times n}$, parameters $\lambda>0$, $\mu^0>0$, $\gamma>1$.\\
{\bfseries Initialize:} $J=I\in \mathbf{\mathcal{R}}^{n\times n}$, $E=0$, $Y_1=Y_2=0$.\\
  {\bfseries REPEAT}
\begin{algorithmic}[1]
   \STATE Obtain $Z$ through (\ref{solveZ}).
   \STATE Update $J$ as (\ref{solveJ}).
   \STATE Solve $E$ by either (\ref{error2}), (\ref{error1}), or (\ref{error21}) according to $l$.
\STATE Update the multipliers:
\begin{align*}
Y_1^{t+1}&=Y_1^{t}+\mu^t(J^{t+1}-Z^{t+1}),\\
Y_2^{t+1}&=Y_2^{t}+\mu^t(X-XZ^{t+1}-E^{t+1}).
\end{align*}
\STATE Update the parameter $\mu^t$ by $\mu^{t+1}=\gamma\mu^t$.
\end{algorithmic}
\textbf{ UNTIL} {stopping criterion is met.}
\end{algorithm}

The complete procedure for solving problem (\ref{ourproblem}) is summarized in Algorithm 1. Since our objective function is nonconvex, it is difficult to give a rigorous mathematical proof for convergence to an (local) optimum. As we show in the supplementary material,
our algorithm converges to an accumulation point and this accumulation point is
a stationary point. Our experiments confirm the convergence of the proposed
method. The experimental results are promising,
despite that the solution obtained by the proposed optimization method may be
a local optimum. 
\subsection{Subspace segmentation} 
After we obtain the coefficient matrix $Z^*$, we consider constructing a similarity graph matrix $W$ from it, since postprocessing of the coefficient matrix often improves the clustering performance \cite{elhamifar2013sparse}. 
Using the angular information based technique in \cite{liu2013robust}, we define $\widetilde{U}=U(\Sigma)^{1/2}$, where $U$ and $\Sigma$ are from the skinny SVD of $Z^*=U\Sigma V^T$. Inspired by \cite{lauer2009spectral}, we define $W$ as:
\begin{equation}
\label{graphmatrix}
W_{ij}=(\frac{\widetilde{u}_i^T\widetilde{u}_j}{\left\|\widetilde{u}_i\right\|_2\left\|\widetilde{u}_j\right\|_2})^{2\phi},
\end{equation}
where $\widetilde{u}_i$ and $\widetilde{u}_j$ stand for the $i$-th and $j$-th columns of $\widetilde{U}$, and $\phi \in {\mathcal{N^*}}$ tunes the sharpness of the affinity between two points. However, an excessively large $\phi$ would break affinities between points of the same group. $\phi=2$ is used in our experiments, and thus we have the same post-processing procedure as LRR\footnote{As we confirmed with an author of \cite{liu2013robust}, the power 2 of equation (12) in \cite{liu2013robust} is a typo, which should be 4. }. 
After obtaining $W$, we directly utilize NCuts to cluster the samples. %Algorithm 2 outlines the complete subspace clustering steps of the proposed method.
\begin{figure}[h]
\centering
\resizebox{.4\textwidth}{!}{
\includegraphics[width=\columnwidth]{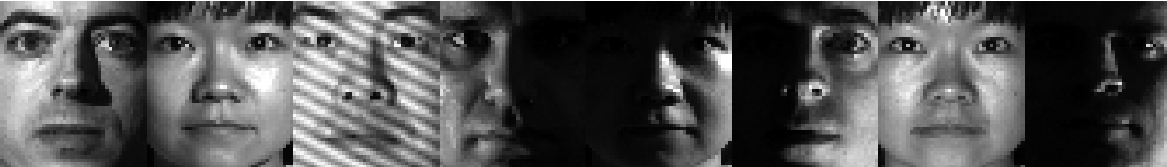}
}
\caption{Sample face images in Extended Yale B.}
\label{img:sampleimage}
\end{figure}

% Note that the IEEE does not put floats in the very first column
% - or typically anywhere on the first page for that matter. Also,
% in-text middle ("here") positioning is typically not used, but it
% is allowed and encouraged for Computer Society conferences (but
% not Computer Society journals). Most IEEE journals/conferences use
% top floats exclusively. 
% Note that, LaTeX2e, unlike IEEE journals/conferences, places
% footnotes above bottom floats. This can be corrected via the
% \fnbelowfloat command of the stfloats package.
\section{Experiment}
In this section, we apply CLAR to subspace clustering on two benchmark databases: the Extended Yale B database (EYaleB) \cite{lee2005acquiring} and the Hopkins 155 motion database \cite{tron2007benchmark}.  CLAR is compared with the state-of-the-art subspace clustering algorithms: SSC, LSA, LRR, and LRSC. The segmentation error rate is used to evaluate the subspace clustering performance, which is defined to be the percentage of erroneously clustered samples versus the total number of samples in the data set being considered. The parameters are tuned to achieve the best performance. In general, when the corruptions or noise are slight, the value of $\lambda$ should be relatively large. For our two experiments, $\lambda =3\times 10^{-4}$ and $67$ are used. $\gamma$ influences the convergence speed, and we adopt $\gamma=1.1$ as often done in literature. For fair comparison, we follow experimental settings in \cite{elhamifar2013sparse}. We stop the program when a maximum of 100 iterations or a relative difference of $10^{-5}$ is reached. The experiments are implemented on Intel Core i5 2.3GHz MacBook Pro 2011 with 4G memory. The code is available at: https://github.com/sckangz/logdet.   

\subsection{Face clustering}
EYaleB consists of 2,414 frontal face images of 38 individuals under 64 lighting conditions. The task is to cluster these images into their individual subspaces by identity. EYaleB is challenging for subspace clustering due to large noise or corruptions, which can be seen from sample images in Figure \ref{img:sampleimage}. As \cite{elhamifar2013sparse}, we model noise with $\|E\|_1$. Each image is resized to a 2016-dimensional vector. We divide the 38 subjects into four groups, i.e., 1 to 10, 11 to 20, 21 to 30, and 31 to 38, and consider all choices of $\{2, 3, 5, 8\}$ for each group and all choices of $n=10$ in the first three groups. There will be $\{163, 416, 812, 136, 3\}$ datasets for each n, respectively.
\begin{figure}[h]
\begin{center}
\resizebox{.55\columnwidth}{!}{
\begin{tabular}{r c l }

\includegraphics[width=.28\columnwidth]{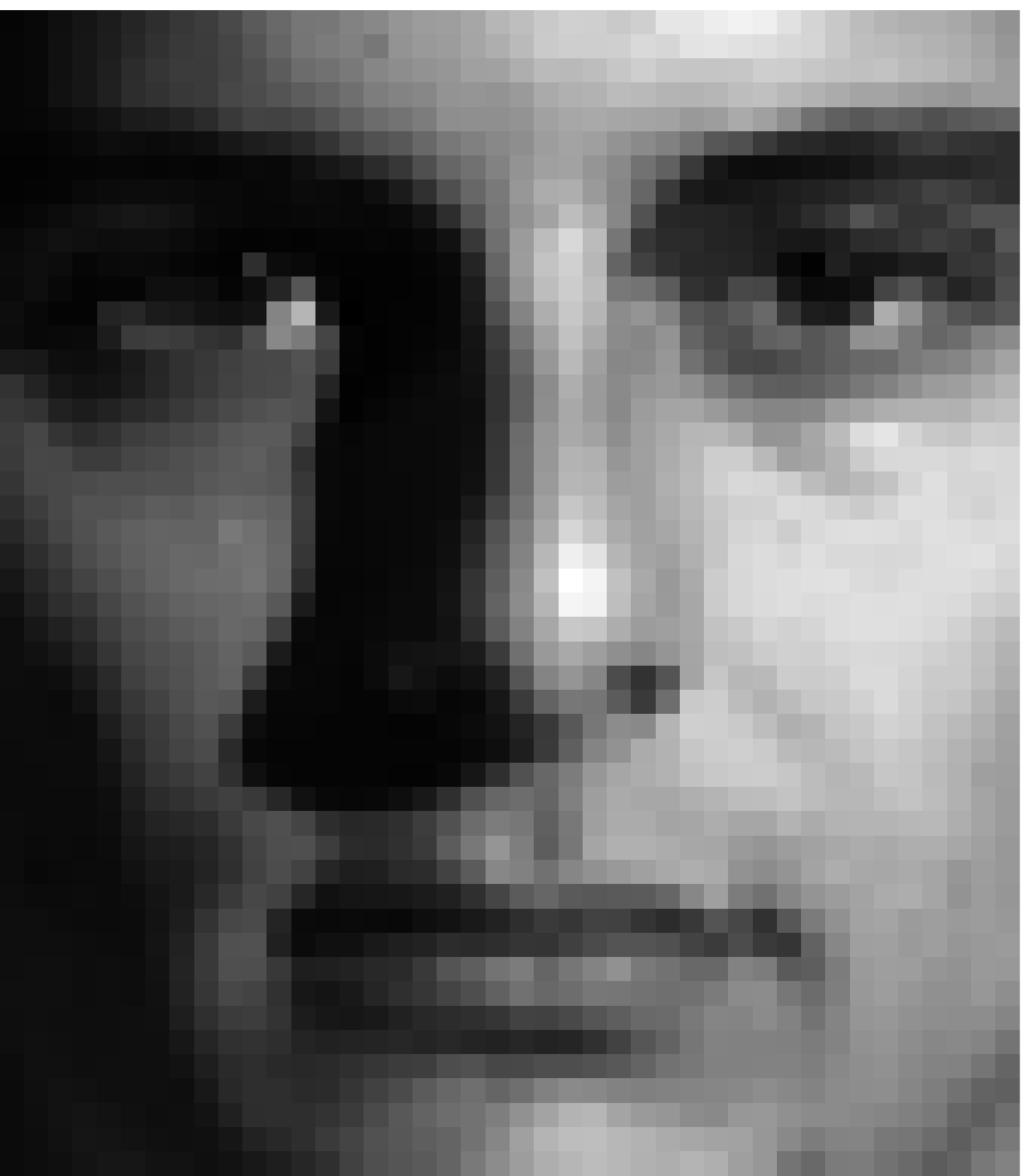}&
\includegraphics[width=.28\columnwidth]{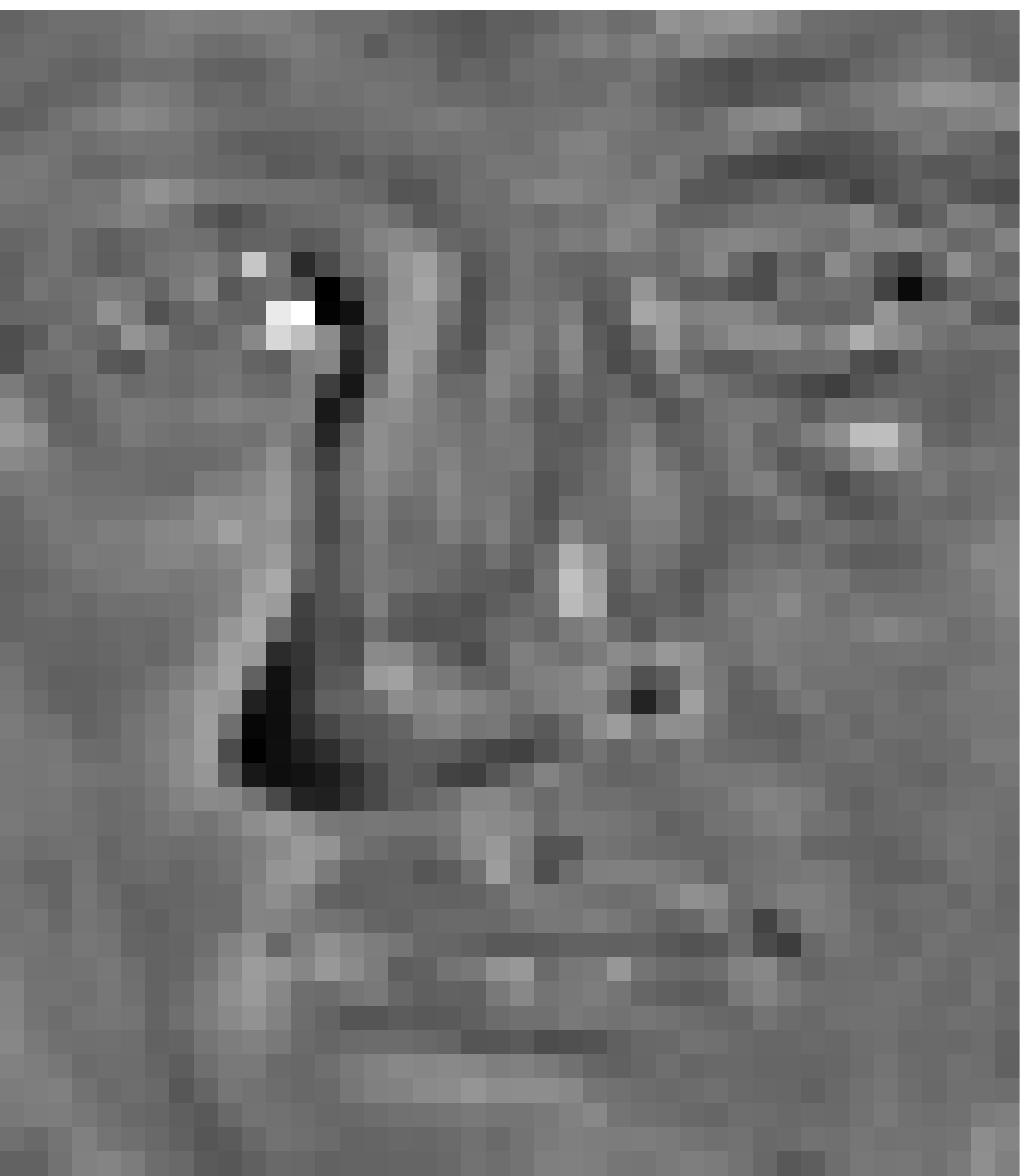}&
\includegraphics[width=.28\columnwidth]{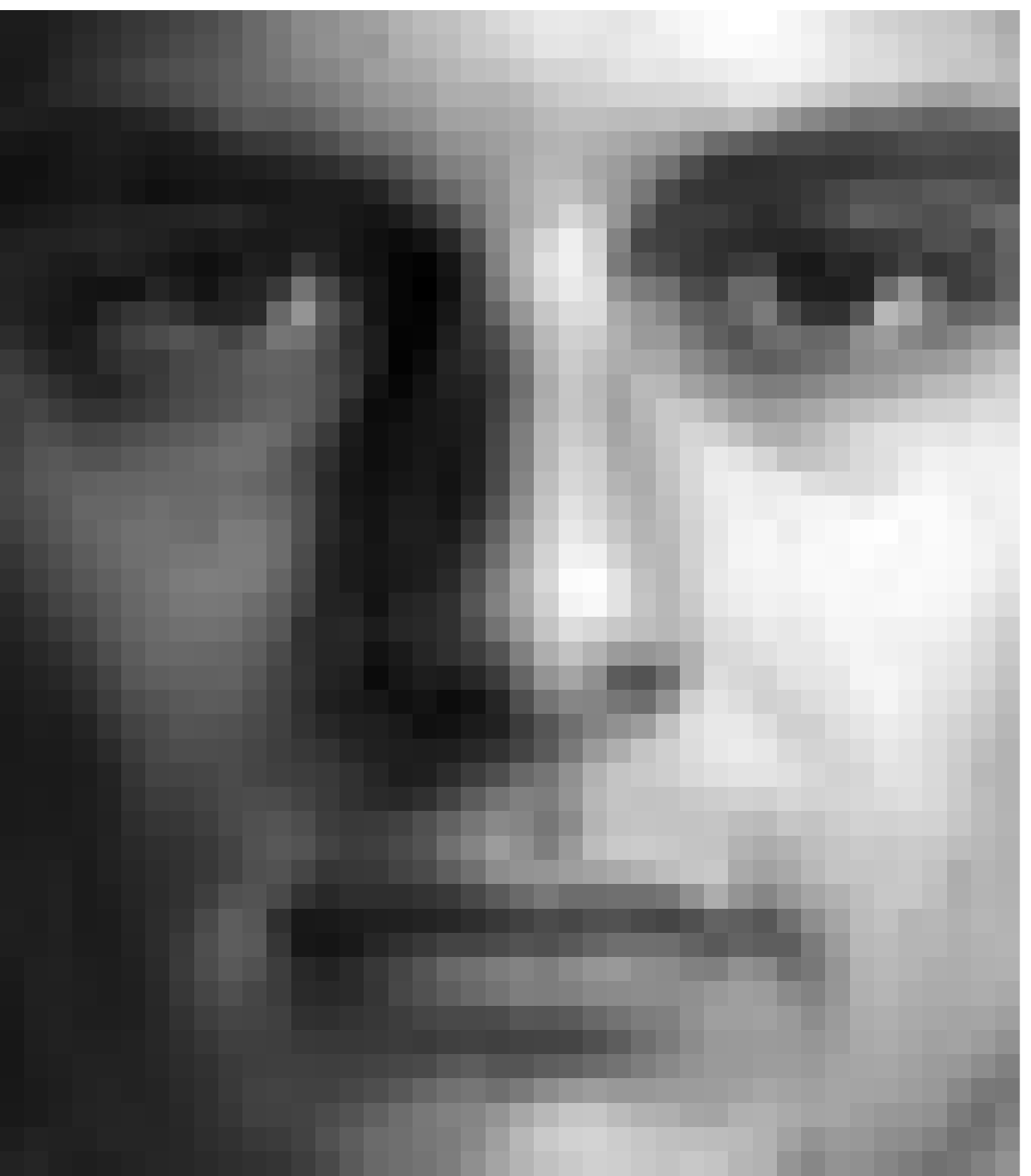}\\
\includegraphics[width=.28\columnwidth]{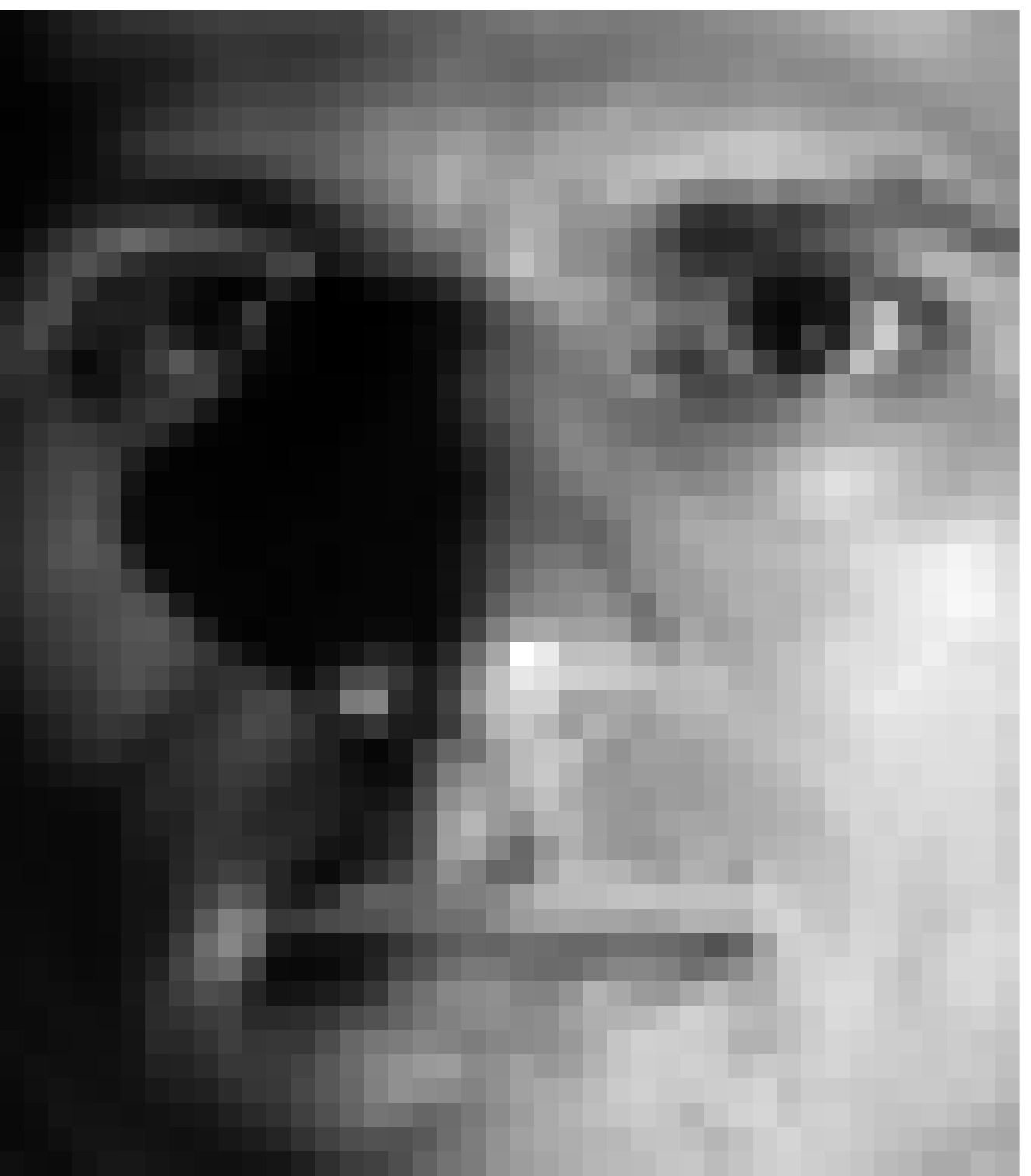}&
\includegraphics[width=.28\columnwidth]{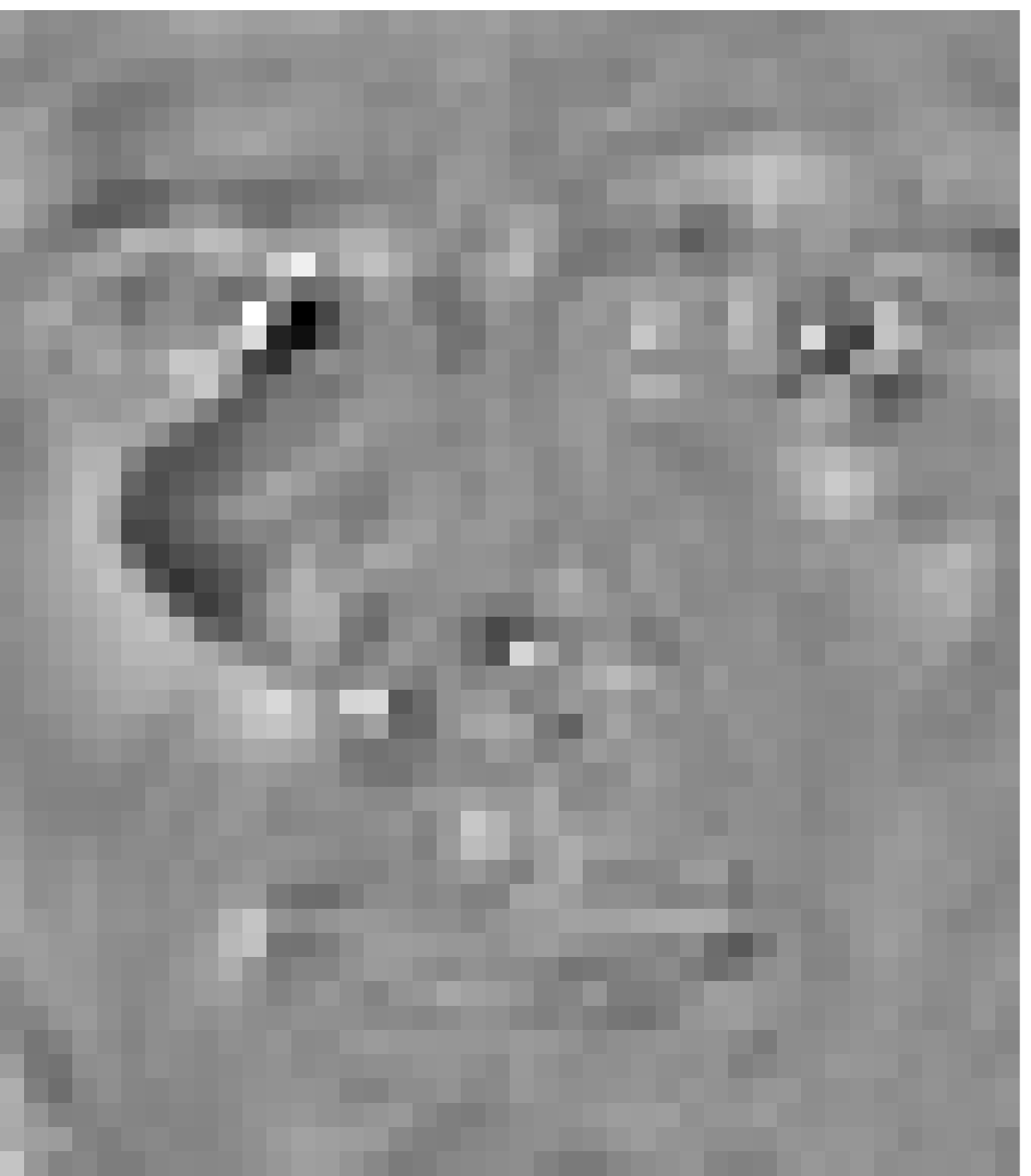}&
\includegraphics[width=.28\columnwidth]{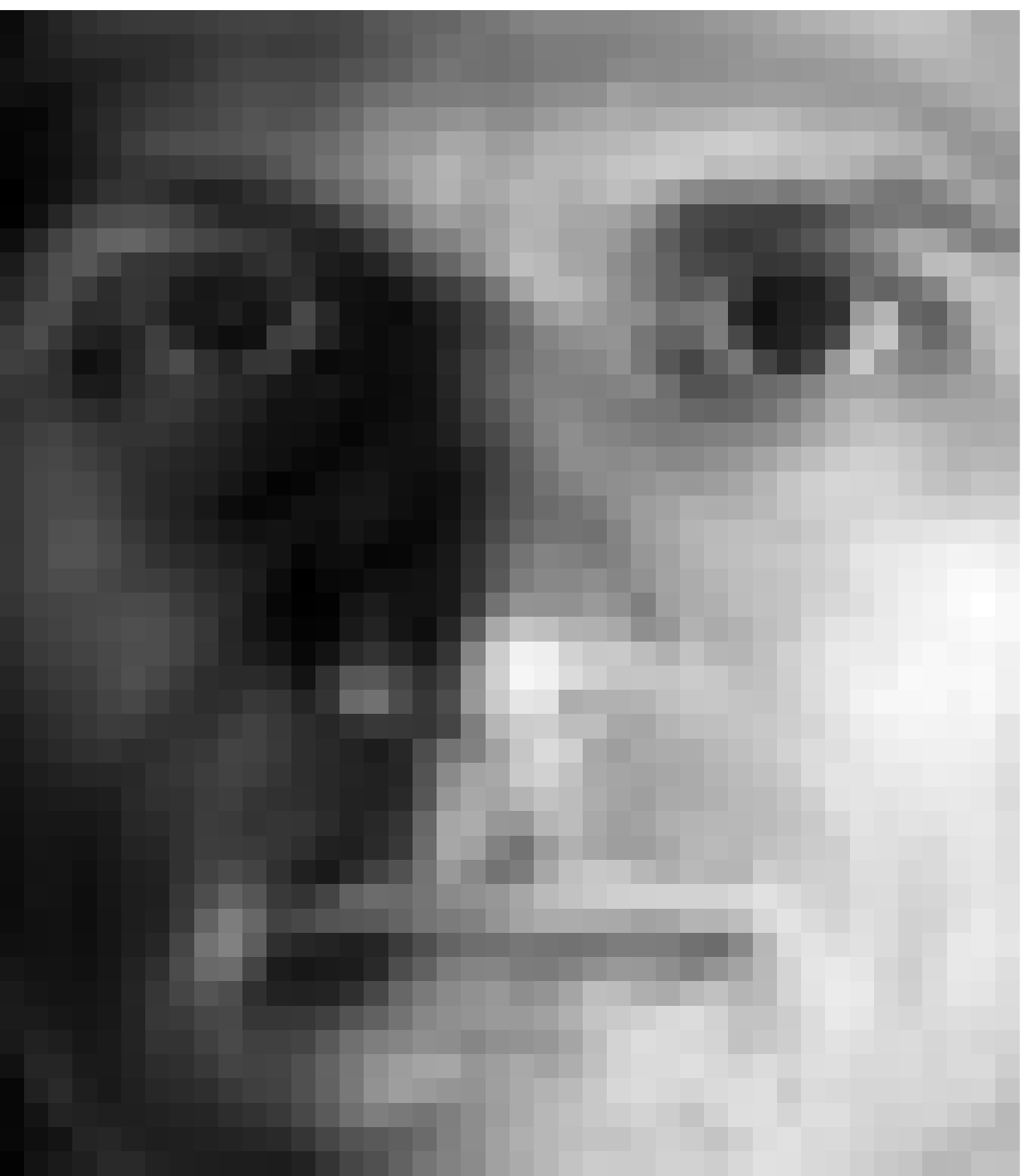}
\end{tabular}}
\caption{ Examples of recovery results of face images. The three columns from left to right are the original image ($X$), the error matrix ($E$) and the recovered image ($XZ$), respectively.}
\label{fig:recoverimage}
\end{center}
\end{figure}

Mean and median error rates for the datasets corresponding to each $n$ are reported in Table~\ref{tab:face}. It can be seen that CLAR outperforms the other methods significantly. As more subjects are involved, the error rate of  CLAR remains at a low level, while those of other methods increase drastically. In particular, in the most challenging case of 10 subjects, the mean clustering error rate of  CLAR is 3.85$\%$, which improves by 7.09$\%$ compared to the best result provided by SSC. This implies that our method is robust to in-sample outliers. In Table~\ref{tab:face}, we also observe that the clustering error rates of LSA are much larger than other methods. This is potentially because LSA is based on MSE, which is heavily influenced by outliers. In addition, the advantage of our method is much more significant with respect to other low-rank representation based algorithms such as LRR and LRSC; for example, there is 11$\%$ and 19$\%$ improvement over LRR in the cases of 8 and 10 subjects, respectively. This verifies the importance of good rank approximation. 

Figure~\ref{fig:recoverimage} shows some recovery results from the 10-subject clustering scenario. As we can see, the error term $E$ is indeed sparse and it helps remove the shadows.
\begin{table}[ht]
\caption{Clustering error rate (\%) on the EYaleB dataset.}
\label{tab:face}
\begin{center}
\begin{small}
\begin{sc}
\resizebox{.3\textwidth}{!}{
\begin{tabular}{llllll}
\multicolumn{1}{c}{\bf METHOD}  &\multicolumn{1}{c}{\bf LRR} &\multicolumn{1}{c}{\bf SSC}  &\multicolumn{1}{c}{\bf LSA} &\multicolumn{1}{c}{\bf LRSC}  &\multicolumn{1}{c}{\bf  CLAR}\\
\hline \\
2 Subjects &&&&&\\
 Mean&2.54&1.86&32.80&5.32&\textbf{1.27} \\
Median&0.78&\textbf{0.00}&47.66&4.69&0.78 \\
\hline
3 Subjects &&&&&\\
 Mean&4.21&3.10&52.29&8.47&\textbf{1.92} \\
Median&2.60&\textbf{1.04}&50.00&7.81&1.56 \\
\hline
5 Subjects &&&&&\\
Mean&6.90&4.31&58.02&12.24&\textbf{2.64} \\
Median&5.63&2.50&56.87&11.25&\textbf{2.19} \\
\hline
8 Subjects &&&&&\\
 Mean&14.34&5.85&59.19&23.72&\textbf{3.36} \\
Median&10.06&4.49&58.59&28.03&\textbf{3.03} \\
\hline
10 Subjects &&&&&\\
Mean&22.92&10.94&60.42&30.36&\textbf{3.85} \\
Median&23.59&5.63&57.50&28.75&\textbf{3.44}  \\
\hline
\end{tabular}}
\end{sc}
\end{small}
\end{center}
\end{table}
\begin{figure}[h]
\begin{center}
\resizebox{.4\textwidth}{!}{
\includegraphics[scale=0.11]{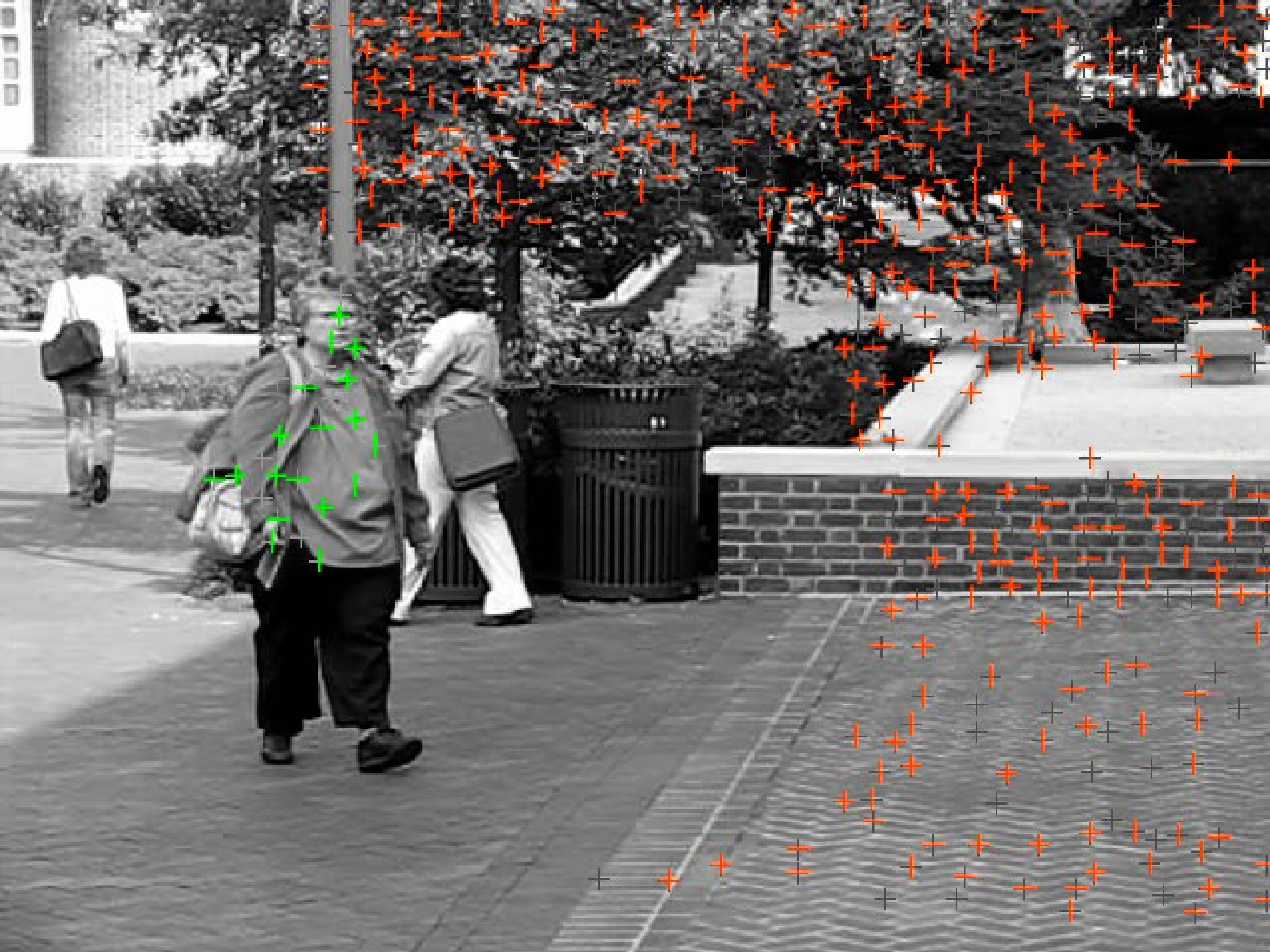}
\includegraphics[scale=0.11]{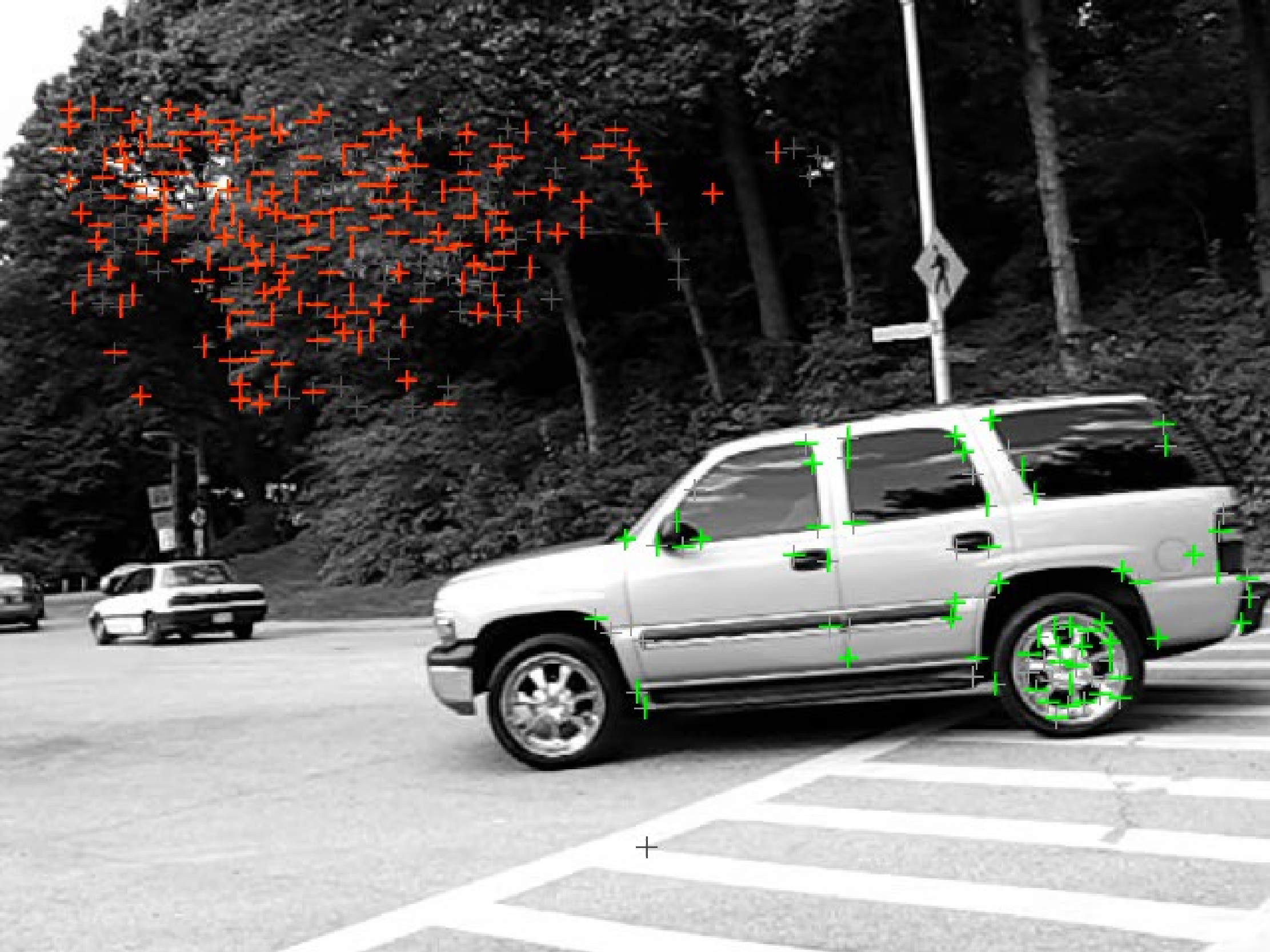}
}
\caption{Sample images in Hopkins 155 database. Trackers are denoted by different colors. }
\label{motionimage} 
%}
\end{center}
\end{figure}
\subsection{Motion segmentation}
In this subsection, we evaluate the robustness of CLAR for motion segmentation problem, which is an important step in video sequences analysis. Given multiple image frames of a dynamic scene, motion segmentation is to cluster the points in those views into different motions undertaken by the moving objects. Hopkins 155 motion database contains 155 video sequences along with features extracted and tracked in all frames for each sequence. 
%Each sequence has 39$\sim$550 data points drawn from two or three motions. There are 120 two motions sequences and 35 three motions. Hopkins 155 database consists of three categories of motions: checkerboard, which has 104 sequences of indoor scenes  taken with a handheld camera under controlled conditions; traffic, which has 38 sequences of outdoor traffic scenes taken by a moving handheld camera; articulated/non-rigid, which has 13 sequences displaying motions constrained by joints, head and face motions, people walking, etc. 
Since the trajectories associated with each motion reside in a distinct affine subspace of dimension $d\leq 3$%\cite{ma2008estimation}
, every motion corresponds to a subspace. Figure \ref{motionimage} gives some sample images. 
$\|E\|_F^2$ is applied to model the noise. 
\begin{table}[ht]
\caption{Segmentation error rate (\%) and mean computational time (s) on the Hopkins 155 dataset.}
\label{tab:motion}
\begin{center}
\begin{small}
\begin{sc}
\resizebox{.28\textwidth}{!}{
\begin{tabular}{llllll}
\multicolumn{1}{c}{\bf METHOD}  &\multicolumn{1}{c}{\bf LRR} &\multicolumn{1}{c}{\bf SSC}  &\multicolumn{1}{c}{\bf LSA} &\multicolumn{1}{c}{\bf LRSC}  &\multicolumn{1}{c}{\bf  CLAR}\\
\hline \\
2 Motions&&&&&\\
 Mean&2.13&1.52&4.23&3.69&\textbf{1.32}  \\
Median&\textbf{0.00}&\textbf{0.00}&0.56&0.29&\textbf{0.00} \\
\hline 
3 Motions &&&&&\\
Mean&4.03&4.40&7.02&7.69&\textbf{2.60}  \\
Median&1.43&0.56&1.45&3.80&\textbf{0.51}  \\
\hline 
All &&&&&\\
Mean&2.56&2.18&4.86&4.59&\textbf{1.61} \\
Median&\textbf{0.00}&\textbf{0.00}&0.89&0.60&\textbf{0.00}  \\
\hline
Average Time&6.44&5.09&17.17&0.70&3.80 \\
\hline
\end{tabular}}
\end{sc}
\end{small}
\end{center}
\end{table}
\begin{figure}[h]
\begin{center}
\includegraphics[width=\columnwidth,height=2.8cm,keepaspectratio]{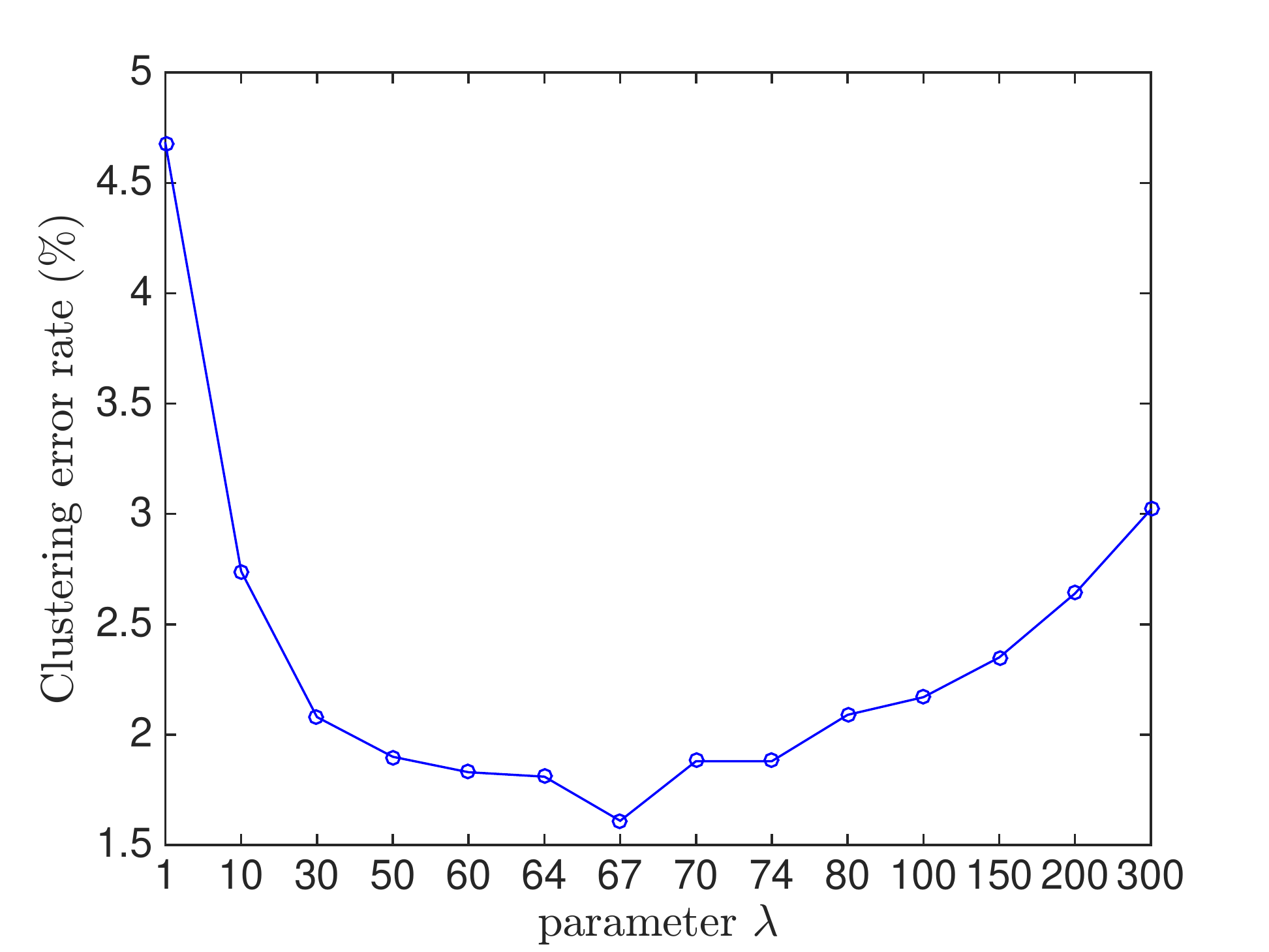}
\caption{The influence of the parameter $\lambda$ of CLAR on all 155 sequences of Hopkins 155.}
\label{parasense}
\end{center}
\end{figure}
Table \ref{tab:motion} shows the clustering results on the Hopkins 155 dataset. CLAR achieves the best results in all cases. Specifically, the average clustering error rate is 1.32$\%$ for two motions and 2.60$\%$ for three motions. We also show the computational time in Table \ref{tab:motion}. As we can see, our computational time is less than LRR, SSC and LSA, though more than LRSC. Figure \ref{parasense} demonstrates the sensitivity of our algorithm to $\lambda$. It shows that the performance of CLAR is quite stable while $\lambda$ varies in a pretty large range. We also test $\gamma$ with values 1.05 and 1.2 which do not give much difference in error rate. Since our problem is nonconvex, we repeat the experiments using different random initializations and we can still get similar results after tuning the parameters. Thus, CLAR appears quite insensitive to initilizations.

\section{Conclusion}
In this paper, we study the matrix rank minimization problem with log-determinant approximation. This surrogate can better approximate the rank function. As an application, we study its use for the robust subspace clustering problem. A minimization algorithm, based on a type of augmented Lagrangian multipliers method, is developed to optimize the associated nonconvex objective function. Extensive experiments on the face clustering and motion segmentation demonstrate the effectiveness and robustness of the proposed method, which shows superior performance when compared to the state-of-the-art subspace clustering methods. %First, it would be interesting to test log-determinant rank approximation in other applications, e.g, matrix completion.

% if have a single appendix:
%\appendix[Proof of the Zonklar Equations]
% or
%\appendix  % for no appendix heading
% do not use \section anymore after \appendix, only \section*
% is possibly needed

% use appendices with more than one appendix
% then use \section to start each appendix
% you must declare a \section before using any
% \subsection or using \label (\appendices by itself
% starts a section numbered zero.)
%

%
%\appendices
%\section{Proof of the First Zonklar Equation}
%Appendix one text goes here.
%
%% you can choose not to have a title for an appendix
%% if you want by leaving the argument blank
%\section{}
%Appendix two text goes here.
%
%
%% use section* for acknowledgment
\section*{Acknowledgment}
This work is supported by US
National Science Foundation grants IIS 1218712.
%
%The authors would like to thank...

% Can use something like this to put references on a page
% by themselves when using endfloat and the captionsoff option.
\ifCLASSOPTIONcaptionsoff
  \newpage
\fi

% trigger a \newpage just before the given reference
% number - used to balance the columns on the last page
% adjust value as needed - may need to be readjusted if
% the document is modified later
%\IEEEtriggeratref{8}
% The "triggered" command can be changed if desired:
%\IEEEtriggercmd{\enlargethispage{-5in}}

% references section

% can use a bibliography generated by BibTeX as a .bbl file
% BibTeX documentation can be easily obtained at:
% http://www.ctan.org/tex-archive/biblio/bibtex/contrib/doc/
% The IEEEtran BibTeX style support page is at:
% http://www.michaelshell.org/tex/ieeetran/bibtex/
%\bibliographystyle{IEEEtran}
% argument is your BibTeX string definitions and bibliography database(s)
%\bibliography{IEEEabrv,../bib/paper}
%
% <OR> manually copy in the resultant .bbl file
% set second argument of \begin to the number of references
% (used to reserve space for the reference number labels box)

% biography section
% 
% If you have an EPS/PDF photo (graphicx package needed) extra braces are
% needed around the contents of the optional argument to biography to prevent
% the LaTeX parser from getting confused when it sees the complicated
% \includegraphics command within an optional argument. (You could create
% your own custom macro containing the \includegraphics command to make things
% simpler here.)
%\begin{IEEEbiography}[{\includegraphics[width=1in,height=1.25in,clip,keepaspectratio]{mshell}}]{Michael Shell}
% or if you just want to reserve a space for a photo:

% insert where needed to balance the two columns on the last page with
% biographies
%\newpage

\bibliographystyle{IEEEtran}
%\bibliography{scref}

\end{document}